\documentclass[letterpaper]{article} 
\usepackage[preprint]{aaai2027}  
\usepackage[hyphens]{url}  
\usepackage{graphicx} 
\usepackage{natbib}  
\usepackage{caption} 
\usepackage{algorithm}
\usepackage{algorithmic}
\usepackage{amsmath}
\usepackage{amssymb}
\usepackage{bm}
\usepackage{booktabs}

\newcommand{\CH}{\mathbf{C}_{\!H}}
\newcommand{\CHc}[1]{C_{H,#1}}
\newcommand{\bw}{\mathbf{w}}
\newcommand{\balpha}{\bm{\alpha}}
\newcommand{\bx}{\mathbf{x}}
\newcommand{\real}{\mathbb{R}}
\newcommand{\Ev}{E_{\mathrm{v}}}

\title{CertMix: Certified, Data-Efficient Metamaterial Design by Affine Mixing of Aligned Neural-Implicit Weight Spaces}

\author{
    Yifan Wang
}
\affiliations{
    Department of Mechanical Engineering, McGill University\\
    Montreal, QC, H3A 2T7, Canada\\
    yifan.wang18@mail.mcgill.ca
}

\begin{document}

\maketitle

\begin{abstract}
Inverse design of mechanical metamaterials---finding a periodic unit cell whose
\emph{homogenized} elastic properties meet a prescribed target---is central to
architected-material engineering, yet current learning-based methods are
data-hungry, confined to interpolation, and provide no guarantee that the
generated design meets specification. We introduce \textbf{CertMix}, a
data-efficient framework that represents each exemplar unit cell as a small
periodic neural implicit field (a SIREN signed-distance decoder) overfit from a
\emph{shared anchor}, so that the exemplar weight vectors become aligned and
directly comparable. Our key observation is that, in this aligned weight space,
the homogenized elasticity tensor is \emph{approximately linear} in the mixing
coefficients; consequently, designing a cell with target properties reduces to a
small constrained \emph{affine mixing} problem solved with a differentiable
periodic homogenizer in the loop. Allowing negative coefficients enables
\emph{extrapolation} beyond the exemplar range, a linearity-mismatch trust region
keeps blends valid, and a split-conformal calibration converts the mismatch signal
into a distribution-free certificate on the achieved-property error. From as few
as \textbf{50} exemplars, CertMix attains a scaled property error of
\textbf{$10^{-4}$}, roughly two to three orders of magnitude below conditional
generative baselines trained on 1000 cells; it remains accurate far outside the
exemplar range, is $57\times$ faster than per-target topology optimization while
avoiding checkerboards and enclosed voids, and extends to spatially graded fields,
$3$D triply-periodic surfaces, and a certified running-shoe midsole application.
\end{abstract}

\section{Introduction}

A mechanical metamaterial derives its effective behaviour from the geometry of a
periodic \emph{unit cell} rather than its base chemistry: by patterning a solid
into a lattice one can realize a target stiffness, a prescribed anisotropy, or a
counter-intuitive negative Poisson's ratio (auxetic response)
\citep{lakes1987auxetic,bertoldi2017metamaterials,surjadi2019metamaterials}. The
practical bottleneck is the \emph{inverse} problem: given a desired homogenized
elasticity tensor $\CH^{*}$, find a manufacturable unit-cell geometry that
realizes it.

Two families of methods dominate. \emph{Topology optimization} (TO) solves a
PDE-constrained optimization per target
\citep{bendsoe1988topopt,sigmund2001topopt99,andreassen2011topopt88}, including the
inverse-homogenization formulation that tailors a periodic cell to prescribed
constitutive parameters \citep{sigmund1994inverse}. TO is accurate but expensive
(a full solve per target) and frequently returns checkerboards or enclosed voids
that are not manufacturable without heavy filtering
\citep{gaynor2016overhang,langelaar2016selfsupport}. \emph{Learned generative
models}---conditional VAEs, GANs, and diffusion models---amortize design across
targets \citep{wang2022ihgan,zheng2023truss,bastek2023videodiffusion,zheng2025diffumeta},
but they typically require $10^3$--$10^4$ simulated cells, are confined to the
convex hull of their training distribution, and offer no guarantee that a
generated cell meets its target once measured.

\begin{figure*}[t]
\centering
\includegraphics[width=0.92\textwidth]{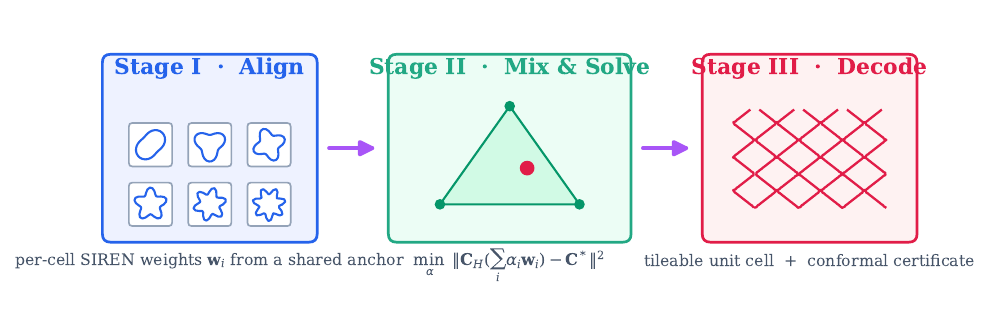}
\caption{\textbf{CertMix in three stages.} (I)~\emph{Align}: overfit a tiny
periodic SIREN signed-distance decoder to each exemplar cell from one shared
anchor, so the flattened weight vectors $\bw_i$ live in a common space.
(II)~\emph{Mix \& solve}: choose affine coefficients $\balpha$ so that the
differentiable homogenized tensor $\CH(\sum_i\alpha_i\bw_i)$ matches the property
target $\CH^{*}$. (III)~\emph{Decode}: read the zero-level set of the blended
weights into a seamless, tileable unit cell, with a split-conformal certificate on
the achieved error.}
\label{fig:pipeline}
\end{figure*}

We take a different route inspired by recent \emph{weight-space} shape models that
control \emph{geometric} parameters by mixing the weights of aligned neural
implicit decoders \citep{plattner2025lamp,berzins2025shapeweight}. Our central
scientific claim is that, once per-exemplar signed-distance decoders are aligned by
overfitting from a shared anchor, the \emph{physical} homogenized property is
approximately \emph{linear} in the weight-space mixing coefficients. This turns
property-controlled design into a tiny constrained linear-blending problem that we
solve with an \emph{exact, differentiable} periodic homogenizer in the loop
(Fig.~\ref{fig:pipeline}).

\paragraph{Contributions.}
\begin{itemize}
\item We show empirically that PDE-homogenized elastic properties are approximately
linear in the coefficients of an \emph{aligned} neural-implicit weight space, and
that a cheap \emph{linearity-mismatch} signal predicts when a blend is trustworthy
(Spearman $0.855$; failure-flag AUC $0.962$).
\item We formulate property-controlled metamaterial design as a constrained
\emph{affine mixing} problem with a differentiable periodic homogenizer in the
loop, giving exact self-adjoint gradients at no extra solve cost, and a two-stage
(convex-then-affine) solver with a trust region for robust \emph{extrapolation}.
\item We attach a \emph{split-conformal certificate} to the achieved-property
error, yielding a distribution-free coverage guarantee, and add exact
manufacturability checks.
\item We extend the method to spatially graded mixing \emph{fields}, to $3$D
triply-periodic surfaces, and to a certified personalized running-shoe midsole,
and report comprehensive comparisons against retrieval, cVAE, diffusion, learned
weight regression, and per-target topology optimization.
\end{itemize}

\section{Related Work}

\paragraph{Neural implicit shape representations.}
Coordinate networks that map a point to occupancy or signed distance
\citep{park2019deepsdf,mescheder2019occupancy} represent geometry compactly and
continuously. Periodic activations \citep{sitzmann2020siren} and Fourier features
\citep{tancik2020fourier} capture high-frequency detail; we use a period-lifted
SIREN so that decoded cells tile seamlessly by construction. Surveys of neural
fields \citep{xie2022neuralfields} document their breadth.

\paragraph{Weight-space learning.}
Treating a trained network's weights as data underlies hypernetworks
\citep{ha2017hypernetworks}, \emph{functa} \citep{dupont2022functa}, and
architectures equivariant to the permutation symmetries of weight space
\citep{navon2023dws}. Alignment from a shared initialization is what makes naive
weight interpolation meaningful, echoing linear mode connectivity modulo
permutation \citep{ainsworth2023gitrebasin}. LAMP \citep{plattner2025lamp} and
related work \citep{berzins2025shapeweight} mix aligned SDF weights to control
\emph{geometric} macro-parameters, validated by a \emph{learned} surrogate.
CertMix differs by controlling \emph{physical} homogenized properties through an
exact PDE solver in the loop, by adding spatially graded fields, and by replacing
heuristic safety scores with a conformal certificate.

\paragraph{Inverse design of metamaterials.}
Numerical homogenization \citep{andreassen2014homogenization} and
inverse-homogenization TO \citep{sigmund1994inverse} are the classical tools.
Learned inverse designers include IH-GAN for cellular structures
\citep{wang2022ihgan}, generative truss-metamaterial models
\citep{zheng2023truss}, and diffusion-based designers
\citep{bastek2023videodiffusion,zheng2025diffumeta}. These require large labeled
datasets and interpolate; CertMix is data-efficient, extrapolates, and certifies.

\paragraph{Conformal prediction.}
Split-conformal methods \citep{vovk2005algorithmic,lei2018distribution,angelopoulos2021gentle}
provide finite-sample, distribution-free coverage. We use them to bound the
achieved-property error, turning a heuristic validity signal into a guarantee.

\section{Method}

\subsection{Aligned neural-implicit exemplars}

Each exemplar unit cell $k$ is represented by a periodic SIREN
$s_{\bw}:[0,1]^2\!\to\!\real$ with weights $\bw\in\real^{D}$ that predicts a signed
distance (solid where $s<0$). Periodicity is enforced by lifting the input
coordinate before the first layer,
\begin{equation}
\gamma(\bx) = \bigl[\cos 2\pi x,\ \sin 2\pi x,\ \cos 2\pi y,\ \sin 2\pi y\bigr],
\label{eq:lift}
\end{equation}
so that any decoded field tiles seamlessly---a property the graded-structure stage
relies on. We overfit every exemplar \emph{from one shared anchor}
initialization $\bw^{0}$, optionally with a proximal pull
$\rho\|\bw-\bw^{0}\|_2^2$. This aligns the weight vectors $\{\bw_i\}_{i=1}^{N}$ so
that they occupy a common, comparable region of weight space; without this
alignment, weight interpolation is meaningless (Sec.~Ablations confirms a
$400\times$ degradation).

\subsection{Differentiable periodic homogenization}

Given a weight vector $\bw$, we decode a signed-distance field on an $N_H\times
N_H$ grid, convert it to a smooth density by a differentiable Heaviside
\begin{equation}
\rho(\bx) = \sigma\!\bigl(-s_{\bw}(\bx)/\beta\bigr),
\qquad
E_e = E_{\min} + (E_0-E_{\min})\,\rho_e,
\label{eq:density}
\end{equation}
and compute the homogenized plane-stress elasticity tensor $\CH\in\real^{3\times3}$
by periodic homogenization using bilinear quadrilateral elements and the
energy / mutual-energy formulation \citep{andreassen2014homogenization}. With
periodic boundary conditions, the induced displacement fields $\bm{\chi}^{(j)}$ for
the three unit test strains solve $K\bm{\chi}^{(j)} = f^{(j)}$, and
\begin{equation}
\CHc{ij} = \sum_{e} E_e\, q^{(e)}_{ij},
\qquad
q^{(e)}_{ij} = (\bm{u}^{0}_i - \bm{\chi}_i)_e^{\!\top} K_e^{0}\,(\bm{u}^{0}_j - \bm{\chi}_j)_e,
\label{eq:CH}
\end{equation}
where $K_e^{0}$ is the unit-modulus element stiffness and $\bm{u}^{0}_j$ are the
affine unit-strain modes. Because each element modulus enters $K$ \emph{linearly},
the sensitivity is self-adjoint and available in closed form,
\begin{equation}
\frac{\partial \CHc{ij}}{\partial E_e} = q^{(e)}_{ij},
\label{eq:sens}
\end{equation}
requiring \emph{no additional linear solves}. We wrap Eqs.~\eqref{eq:CH}--\eqref{eq:sens}
as a custom autograd operator, making the map $\bw\mapsto\CH$ end-to-end
differentiable. Engineering constants follow from the compliance
$\mathbf{S}=\CH^{-1}$: $E_x = 1/S_{11}$, $E_y = 1/S_{22}$,
$\nu_{xy} = -S_{12}/S_{11}$, and $G = \CHc{33}$. A $3$D trilinear-hexahedral analog
with an algebraic-multigrid solver handles triply-periodic surfaces.

\subsection{Property-controlled affine mixing}

Let $F(\balpha) = \sum_{i=1}^{N}\alpha_i\,\bw_i$ be the blended weight vector.
Property-controlled design solves
\begin{equation}
\begin{aligned}
\min_{\balpha}\quad & \bigl\| \CH\!\bigl(F(\balpha)\bigr) - \CH^{*} \bigr\|_{\mathrm{sc}}^{2}
 \;+\; \lambda_{\mathrm{m}}\,m(\balpha) \;+\; \lambda_{1}\,\|\balpha\|_{1} \\
\text{s.t.}\quad & \textstyle\sum_{i}\alpha_i = 1,
\end{aligned}
\label{eq:mix}
\end{equation}
where $\|\cdot\|_{\mathrm{sc}}$ is a per-property standardized (scaled) $L_2$ norm
over the controlled entries of $\CH$. The affine constraint $\sum_i\alpha_i=1$
keeps blends on the affine hull of the exemplars; permitting $\alpha_i<0$ lets the
design leave the convex hull, which is precisely what enables extrapolation. The
\emph{linearity mismatch}
\begin{equation}
m(\balpha) = \Bigl\| s_{F(\balpha)} - \textstyle\sum_i \alpha_i\, s_{\bw_i} \Bigr\|_1
\label{eq:mismatch}
\end{equation}
measures how far the decoded field of the blended weights departs from the linear
blend of the exemplar fields; it acts as a differentiable \emph{trust region}
keeping the solve inside the locally linear regime.

\paragraph{Two-stage solve.}
We first solve a \emph{convex} restriction with $\balpha=\mathrm{softmax}(\bm z)$
(guaranteed feasible), then \emph{refine} with a free affine parameterization
initialized at the convex optimum under a strong mismatch penalty. Since the affine
feasible set contains the convex one and we keep the better of the two, affine
refinement can only improve on the convex solution. Optimization uses Adam
\citep{kingma2015adam}; each iteration performs one homogenization solve and uses
the closed-form gradient of Eq.~\eqref{eq:sens}.

\subsection{Spatially graded mixing fields}

Replacing the single coefficient vector by a smooth field
$\balpha(\bx) = \mathrm{softmax}\bigl(\mathrm{bilinear}(\mathbf{Z},\bx)\bigr)$ over
macro-grid logits $\mathbf{Z}$ yields \emph{functionally graded} structures: each
macro cell is homogenized with its centre coefficients to match a prescribed
property field, and a smoothness penalty on $\mathbf{Z}$ regularizes the field.
Because every macro cell is decoded from the same aligned, periodic network, the
implicit fields agree at shared boundaries, so the global structure is continuous
by construction---interfaces are $9$--$23\times$ smoother than tiling distinct
cells.

\subsection{Split-conformal property certificate}

To turn validity into a guarantee we calibrate a split-conformal bound
\citep{lei2018distribution,angelopoulos2021gentle}. On a calibration set we fit a
log-linear scale model $g(\cdot)$ of the achieved error from cheap features
(mismatch, hull distance, $\|\balpha\|_1$), form nonconformity scores
$r = \mathrm{err}/g$, and take the quantile
\begin{equation}
\hat q = \mathrm{Quantile}_{\lceil (n+1)(1-\delta)\rceil / n}\bigl(\{r_i\}\bigr).
\label{eq:conformal}
\end{equation}
At test time the certified error bound for a new design is $\hat q\, g(\cdot)$,
which holds with probability $\ge 1-\delta$ under exchangeability.

\begin{algorithm}[t]
\caption{CertMix property-controlled design}
\label{alg:certmix}
\textbf{Input}: aligned bank $\{\bw_i\}$, target $\CH^{*}$, calibrated $(\hat q, g)$\\
\textbf{Output}: unit-cell weights $\bw^{\star}$, certified error bound $b$
\begin{algorithmic}[1]
\STATE $\balpha \gets$ convex solve of Eq.~\eqref{eq:mix} ($\mathrm{softmax}$)
\STATE $\balpha \gets$ affine refine from $\balpha$ under trust region $\lambda_{\mathrm{m}}$
\STATE $\bw^{\star} \gets \sum_i \alpha_i \bw_i$;\ \ evaluate $\CH(\bw^\star)$, $m(\balpha)$
\STATE $b \gets \hat q \cdot g(m,\ \text{hull dist},\ \|\balpha\|_1)$
\STATE \textbf{return} $\bw^{\star}$, $b$
\end{algorithmic}
\end{algorithm}

\section{Experiments}

\paragraph{Setup.}
Unless noted, the exemplar bank is procedurally generated from five $2$D periodic
families (grid, cross-brace, perforated plate, honeycomb spanning
conventional$\leftrightarrow$re-entrant/auxetic, and chiral), each decoded by a
$3$-layer periodic SIREN. All methods---ours, generative baselines, and TO---are
scored by the \emph{same} exact homogenizer at in-loop resolution $N_H=64$,
verified for resolution independence at $N_H=128$. The controlled property set is
$\{E_x,\nu_{xy}\}$ (all of $[E_x,E_y,\nu_{xy},G]$ reported); error is the
per-property standardized $L_2$ distance to target. Experiments run on a CPU-only
laptop. The homogenizer passes five validation gates (analytic solid tensor to
$10^{-15}$; rank-1 laminate matching the Backus average to $1.4\%$; auxetic
honeycomb giving $\nu_{xy}<0$; autograd vs.\ finite-difference agreement to
$<10^{-7}$; $3$D isotropy of a solid and gyroid).

\subsection{Weight-space property linearity}

\begin{figure}[t]
\centering
\includegraphics[width=0.85\columnwidth]{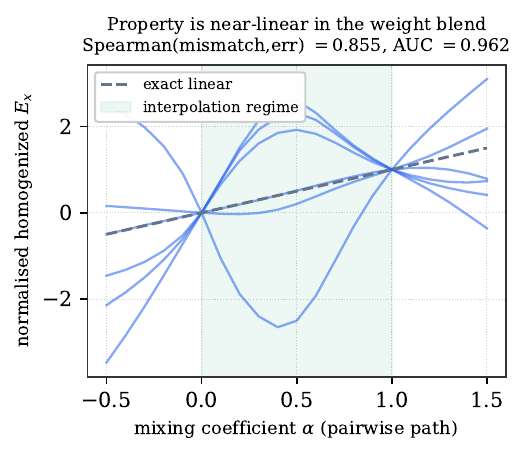}
\caption{Homogenized property along pairwise weight-mixing paths stays near the
exact-linear reference inside $[0,1]$ and bends gently outside. The mismatch signal
predicts true error (Spearman $0.855$, AUC $0.962$).}
\label{fig:linearity}
\end{figure}

Figure~\ref{fig:linearity} plots the homogenized $E_x$ along pairwise mixing paths.
Inside the interpolation regime $\alpha\in[0,1]$ the property tracks the exact
linear reference closely (mean relative error $0.041$), and it bends only mildly
outside it ($0.108$). Over $400$ random blends, the linearity mismatch
$m(\balpha)$ ranks strongly with the true property error (Spearman $0.855$) and
flags large-error blends with AUC $0.962$---comparable to the surrogate-based
safety metric of \citet{plattner2025lamp} but grounded in the real PDE error rather
than learned labels. This validates both the affine formulation and the mismatch
trust region.

\subsection{Data efficiency}

\begin{figure}[t]
\centering
\includegraphics[width=0.86\columnwidth]{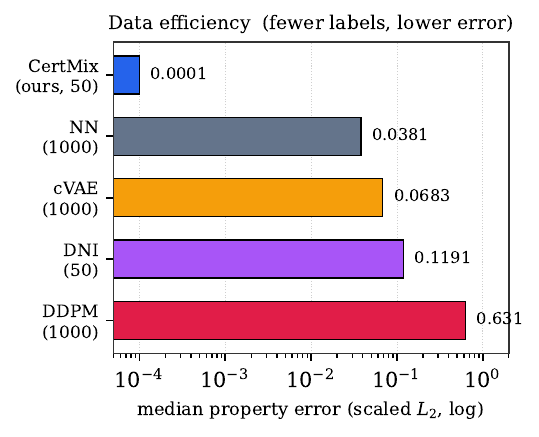}
\caption{Data efficiency: CertMix from $50$ exemplars vs.\ baselines trained on up
to $1000$ cells (log scale).}
\label{fig:eff}
\end{figure}

Table~\ref{tab:main} and Fig.~\ref{fig:eff} compare CertMix (50 exemplars) against
nearest-exemplar retrieval, a conditional VAE \citep{kingma2014vae}, a
classifier-free guided DDPM \citep{ho2020ddpm,sohldickstein2015diffusion}, and a
learned weight-regression network (DNI). Generative baselines receive a generous
best-of-eight sampling budget. CertMix reaches a median scaled error of
$\mathbf{0.0001}$ from $50$ cells---about $680\times$ below the best cVAE trained on
$1000$ cells and three orders of magnitude below the diffusion model.

\begin{table}[t]
\centering
\begin{tabular}{lcc}
\toprule
\textbf{Method} & \textbf{Train cells} & \textbf{Median error} $\downarrow$\\
\midrule
\textbf{CertMix (ours)} & \textbf{50} & \textbf{0.0001}\\
Nearest exemplar & 1000 & 0.0381\\
Conditional VAE & 1000 & 0.0683\\
Learned weights (DNI) & 50 & 0.1191\\
Conditional diffusion & 1000 & 0.6310\\
\bottomrule
\end{tabular}
\caption{Property-target error (scaled $L_2$), scored by the identical homogenizer.
Generative baselines use best-of-eight sampling.}
\label{tab:main}
\end{table}

\subsection{Certified extrapolation}

\begin{figure}[t]
\centering
\includegraphics[width=0.86\columnwidth]{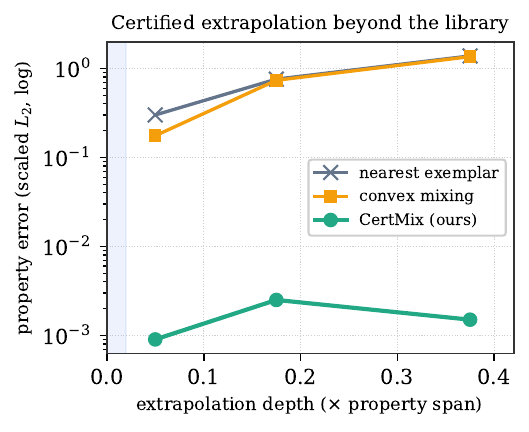}
\caption{Property error vs.\ extrapolation depth. Interpolation-only methods break
down; CertMix stays accurate well beyond the exemplar range.}
\label{fig:extrap}
\end{figure}

We target properties drawn from an \emph{extended} parameter box lying outside the
bank's convex hull, laddered by extrapolation depth (as a fraction of the property
span). Figure~\ref{fig:extrap} shows that convex mixing and nearest-exemplar
retrieval diverge with depth, whereas the two-stage affine solve stays on target.
At $25$--$50\%$ depth CertMix attains error $0.0015$ versus $1.36$ for convex
mixing---roughly a $900\times$ reduction. An ablation confirms the trust region is
essential: raw affine without it reaches a max error of $1.31$, versus $0.026$ for
the two-stage solve.

The split-conformal certificate (Eq.~\eqref{eq:conformal}) achieves $\mathbf{95.8\%}$
empirical coverage at a target of $90\%$, with a failure-flag AUC of $0.795$ and
recall $1.00$ at the chosen tolerance.

\subsection{Functionally graded structures}

With a spatially varying mixing field, CertMix produces continuous graded parts.
For a Poisson-ratio grading (auxetic transition $\nu_{xy}:-0.26\!\to\!+0.83$ with
$E_x$ held fixed) the per-cell property error is $0.0052$ (median) and the interface
seam mismatch is $9\times$ smoother than nearest-exemplar tiling; a global tension
test recovers the target column-$\nu$ profile with Pearson $r=0.997$. A
cross-topology transition (grid$\to$honeycomb) yields $23\times$ smoother seams.

\subsection{Comparison to topology optimization}

\begin{figure}[t]
\centering
\includegraphics[width=0.86\columnwidth]{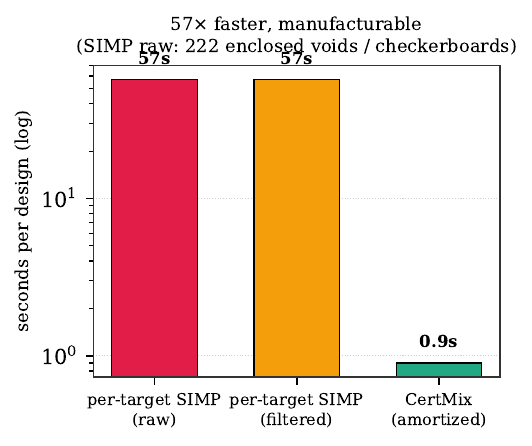}
\caption{CertMix is $57\times$ faster than per-target SIMP and manufacturable by
construction (SIMP-raw: $222$ enclosed voids).}
\label{fig:simp}
\end{figure}

Against per-target inverse-homogenization TO (SIMP), CertMix matches or beats the
converged property error while being amortized: $57\times$ faster per design
(Fig.~\ref{fig:simp}), with $0$ enclosed voids and no minimum-feature violation,
whereas raw SIMP produces checkerboards ($222$ enclosed voids) and filtered SIMP
fails to converge on several targets.

\subsection{Ablations}

\emph{Shared anchor} is decisive: anchored alignment yields median error $0.0013$
versus $0.536$ for independent initialization ($\sim\!400\times$). The trust-region
weight $\lambda_{\mathrm{m}}$ trades property accuracy against mismatch as expected;
in-loop resolution $N_H=64$ suffices for resolution-independent design (verified at
$N_H=128$); and the bank spans $10$--$50$ cells with graceful degradation.

\subsection{Application: certified personalized midsoles}

\begin{figure}[t]
\centering
\includegraphics[width=0.99\columnwidth]{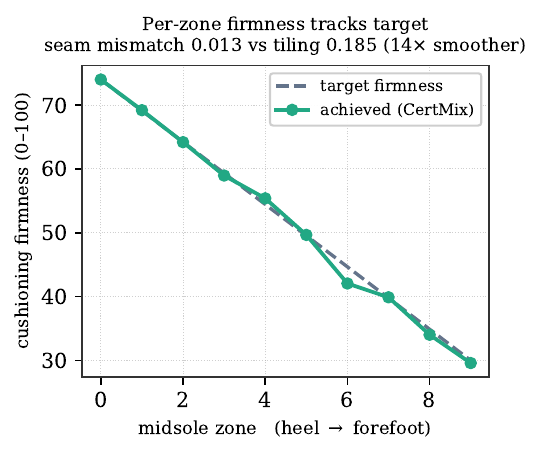}
\caption{Per-zone cushioning firmness of a graded midsole tracks the prescribed
heel$\to$forefoot target; seams are $\sim\!14\times$ smoother than block tiling.}
\label{fig:midsole}
\end{figure}

Re-parameterizing the controlled property as the vertical compression stiffness
$\Ev = \CHc{22}$---the effective modulus footwear science uses for cushioning
firmness \citep{gibson1997cellular}---CertMix designs functionally graded
running-shoe midsoles over a $100$-cell library spanning a $90\times$ firmness range
(Fig.~\ref{fig:midsole}). Personalizing per-zone firmness to runner body weight
attains median error $0.0001$ versus $0.0119$ for the nearest catalog cell; a
graded heel-to-forefoot midsole tracks its target firmness profile with median error
$0.004$ and $14\times$ smoother seams than tiling, and is exported as a printable
STL. For runners requiring support firmer than any catalog pattern, negative
coefficients extrapolate up to $190\%$ beyond the catalog band while the conformal
certificate still holds at the target coverage.

\section{Conclusion}

CertMix casts property-controlled metamaterial design as constrained affine mixing
in an aligned neural-implicit weight space, exploiting the near-linearity of
PDE-homogenized properties in weight-space coordinates. With a differentiable
homogenizer in the loop, a linearity-mismatch trust region, and a split-conformal
certificate, it is data-efficient, extrapolative, fast, manufacturable, and
certified---and extends to graded fields, $3$D surfaces, and a footwear
application. Limitations include the linear-elastic property regime (large-strain
energy return is future work) and the reliance on a procedurally generated exemplar
bank; scaling to richer $3$D families and nonlinear responses are natural next
steps.

\bibliography{references}

\end{document}